\documentclass[letterpaper]{article}
\usepackage[preprint]{aaai2027}
\usepackage[hyphens]{url}
\usepackage{graphicx}
\usepackage{natbib}
\usepackage{caption}
\usepackage{algorithm}
\usepackage{algorithmic}
\usepackage{newfloat}
\usepackage{listings}
\DeclareCaptionStyle{ruled}{labelfont=normalfont,labelsep=colon,strut=off}
\floatstyle{ruled}
\newfloat{listing}{tb}{lst}{}
\floatname{listing}{Listing}

\usepackage{booktabs}
\usepackage{enumitem}
\usepackage[table]{xcolor}
\usepackage{multirow}
\usepackage{tabularx}
\usepackage{amsmath}
\usepackage{amssymb}
\usepackage{mathtools}
\usepackage{amsthm}
\usepackage{tcolorbox}
\tcbuselibrary{breakable}
\usepackage{threeparttable}
\definecolor{pendingred}{RGB}{180,0,0}

\definecolor{verifyblue}{RGB}{0,75,170}

\theoremstyle{plain}

\theoremstyle{definition}

\theoremstyle{remark}

\title{AutoRefine: Compiling Trajectories into Validated Typed Agent Artifacts}

\renewcommand{\thefootnote}{\fnsymbol{footnote}}

\author{
    Libin Qiu\textsuperscript{\rm 1}\thanks{Equal contribution.
    \textsuperscript{\rm 1}Alibaba Group.
    \textsuperscript{\rm 2}Tongji University.
    Correspondence to: Yuhang Ye \textless yeyu.yyh@alibaba-inc.com\textgreater.},
    Zhirong Gao\textsuperscript{\rm 1}\footnotemark[1],
    JunFu Chen\textsuperscript{\rm 1},
    Yuhang Ye\textsuperscript{\rm 1},
    Liangyu Li\textsuperscript{\rm 2},\\
    Wenkai Qiu\textsuperscript{\rm 1},
    Weizhi Huang\textsuperscript{\rm 1},
    Xiaobo Xue\textsuperscript{\rm 1},
    Shuo Tang\textsuperscript{\rm 1}
}
\affiliations{
}

\begin{document}

\maketitle
\renewcommand{\thefootnote}{\arabic{footnote}}
\setcounter{footnote}{0}

\begin{abstract}
Large language model agents repeatedly encounter related tasks, yet systems that learn from trajectories commit every lesson to one predefined artifact form. A local constraint, a reusable procedure, and a delegated objective require different amounts of runtime ownership, so one form either under-specifies the correction or wraps it in execution machinery it does not need. We present \textbf{AutoRefine}, which treats trajectory learning as typed artifact compilation. It contrasts failed against successful trajectories to derive a type-neutral, evidence-linked intervention specification, then compiles that specification into the first Rule, Skill, or bounded Subagent that closes it under a runtime-relative ownership order: the selected schema must own every specified observation, state variable, dependent decision, and completion condition. Validation is stated in the same terms: a type-specific contract gate tests whether the generated object realizes its declared boundary, and a replay gate admits it only when it improves the correction cases linked to its source failures without regression on preservation cases. With GPT-5.6-terra as the shared backbone, AutoRefine records the highest success on ALFWorld, ScienceWorld, TravelPlanner, and SpreadsheetBench, and ties the best result on SkillCraft; on TravelPlanner it reaches 80.56\% success against 50.0\% for the strongest baseline. Removing boundary closure or replay validation costs 15.00 and 16.11 percentage points, the two largest losses among our construction and admission policies. In a longitudinal TravelPlanner stream, the repository holds 89--91\% held-out success after 60 learning tasks with no net loss on previously solved tasks, and frozen repositories improve all 25 evaluated source--target pairs, more within a domain (14.20 points on average) than across domains (6.99). Representation selection and validation therefore act as coupled requirements for reliable capability accumulation.

\noindent\textbf{Code:} \url{https://github.com/AutoRefine/Autorefine}
\end{abstract}

\section{Introduction}

Large language model (LLM) agents now solve long-horizon tasks through reasoning, tool use, and interaction with external environments. Completing a task, however, does not by itself improve a frozen agent: the trajectory is discarded unless its useful evidence becomes persistent external state. Recent work therefore externalizes experience as demonstrations, guidelines, manuals, workflows, playbooks, skills, or executable subagents~\cite{sarukkai2025self,fu2024autoguide,chen2024automanual,wang2025awm,zhang2026ace,yang2026skillopt,zhang2026agentfactory}. These methods share one premise: external artifacts can accumulate capability without updating model parameters.

Most systems, though, optimize inside a representation they fix in advance, such as rules~\cite{chen2024automanual}, skills~\cite{yang2026skillopt,wang2026skillx,lin2026museautosk}, or subagents~\cite{zhang2026agentfactory}. Two coupled challenges follow. \textbf{First, representation adequacy:} a local constraint, a reusable procedure, and a delegated objective require different ownership of observations, state, and decisions, so forcing them into one substrate either under-specifies the intervention or adds execution machinery it does not need. A budget pass/fail check and an exhaustive lodging search differ in exactly this way rather than in wording: only the second needs private search state and its own termination test. A Subagent, in particular, is defined by an explicit interface, isolated state, an independent control loop, and standalone termination, not by prompt length or a specialist role.

\textbf{Second, validation adequacy:} an end-task score alone cannot establish that an artifact implements its claimed execution boundary, nor that repairing its source failures preserves behavior that already succeeds. Because representation determines both execution semantics and failure modes, artifact selection and validation must be designed jointly.

We introduce \textbf{AutoRefine}, which treats trajectory learning as \emph{typed artifact compilation}. Our central design choice is to compile each supported intervention into the first execution boundary that closes its evidence-supported requirements, instead of predicting an artifact type from surface features. AutoRefine contrasts failed against successful trajectories to derive a type-neutral \emph{intervention specification}: when it should fire, what behavior it must change, and which observations, persistent state, dependent decisions, and completion condition it needs, with every field linked to trajectory spans. A heuristic compiler then attempts three schemas in a runtime-relative order of increasing ownership: a \textbf{Rule} for a local decision, a \textbf{Skill} for a parent-executed procedure, and a \textbf{Subagent} for a delegated objective with an independent control loop. It selects the first schema that closes all specified obligations and otherwise defers the candidate. Minimality is relative to that search order and the extracted specification, not a universal expressivity ordering.

Artifact creation is coupled to repository admission. A generated artifact first passes a type-specific \emph{contract gate} that checks whether it realizes the selected execution boundary. A paired \emph{replay gate} then requires improvement on correction cases linked to the source failures without observed regression on preservation cases derived from contrastive successes. Revisions follow the same gated path, so every repository update stays tied to both its behavioral claim and its supporting evidence.

We evaluate AutoRefine on five task benchmarks with a shared GPT-5.6-terra backbone, plus a longitudinal TravelPlanner learning stream and SkillFlow cross-domain transfer. It records the highest success on four benchmarks and ties the best result on SkillCraft, reaching 80.56\% on TravelPlanner against 50.0\% for the strongest baseline; removing boundary-closure checking or the replay gate costs 15.00 and 16.11 percentage points. The longitudinal repository holds 89--91\% held-out success after 60 learning tasks without net degradation, and all 25 transfer pairs improve.

Our contributions are summarized as follows:
\begin{itemize}[itemsep=2pt, topsep=1pt, parsep=0pt]
    \item We formulate trajectory learning as typed artifact compilation: evidence, not a predefined representation, decides whether a lesson becomes a Rule, a Skill, or a bounded Subagent, through first closure in a runtime-relative ownership order.
    \item We develop an evidence-grounded compile-and-check procedure and couple it with representation-specific contract validation and preservation-aware replay for repository admission.
    \item We evaluate end-task performance, representation ablations, repository evolution, and cross-domain transfer, and find that boundary selection and validation contribute separately: removing either causes the largest losses among the construction and admission policies we ablate.
\end{itemize}

\section{Related Work}

\subsection{Externalizing Experience for Frozen Agents}

Recent work treats the non-parametric state around a frozen LLM as an adaptation substrate. Systems abstract experience into reflections, insights, guidelines, and manuals~\cite{shinn2023reflexion,zhao2024expel,fu2024autoguide,chen2024automanual}, accumulate demonstrations~\cite{sarukkai2025self}, induce workflows~\cite{wang2025awm}, or evolve playbooks~\cite{zhang2026ace,wu2026evolver}. Their update policies differ, but each fixes the representation before asking what execution boundary a correction requires. AutoRefine instead derives a type-neutral intervention and compiles it into the first boundary, in a runtime-relative ownership order, that owns its observations, state, decisions, and completion condition.

\subsection{Reusable Skills, Harnesses, and Subagents}

\paragraph{Skill accumulation.}
Executable memories range from program libraries and code-style plans~\cite{wang2023voyager,sun2023adaplanner} to systems that evolve the skill itself: SkillOpt retains bounded edits that improve held-out performance~\cite{yang2026skillopt}, SkillX structures trajectory-derived skills~\cite{wang2026skillx}, MUSE-Autoskill manages their lifecycle~\cite{lin2026museautosk}, and CoEvoSkills generates multi-file packages with a co-evolving verifier~\cite{zhang2026coevoskills}. All of them assume that a reusable correction should be a Skill. AutoRefine addresses the preceding decision: whether the evidence supports a local Rule, a parent-executed Skill, or a delegated Subagent.

\paragraph{Harness evolution.}
Auto-harness methods broaden the editable state beyond skills: Self-Harness admits bounded modifications through regression tests~\cite{zhang2026selfharness}, Agentic Harness Engineering edits heterogeneous files under falsifiable outcome predictions~\cite{lin2026ahe}, and Adaptive Auto-Harness routes among variants for non-stationary streams~\cite{liu2026adaptiveharness}. Where these systems improve or route a broader harness, AutoRefine compiles one supported intervention into its first closed ownership boundary.

\paragraph{Executable subagents.}
AgentFactory is the closest reusable-Subagent precedent, preserving solutions as executable Python subagents~\cite{zhang2026agentfactory}. AutoRefine selects this type only for a closed intermediate objective that requires an independent loop or private state, and its Subagents are separated from Rules and Skills by explicit I/O, isolation, scoped tools, termination, and structured return. Several systems above were 2026 preprints at submission time, so we cite them as contemporaneous design context; only SkillOpt and AgentFactory enter the baseline table.

\subsection{Validation-Gated External Artifacts}

Validation increasingly governs external updates: skill and harness edits are retained through held-out evaluation, regression tests, execution feedback, or learned verifiers~\cite{yang2026skillopt,zhang2026selfharness,lin2026museautosk,zhang2026coevoskills}, while SEVerA guards model calls with formally verified output contracts~\cite{banerjee2026severa}. AutoRefine couples validation to representation: a contract gate checks whether the generated object realizes its ownership boundary, and replay tests source-linked corrections together with contrastive preservation cases. The gates separate structural validity from behavioral utility, since an artifact may satisfy its interface yet regress a previously successful task.

\section{Method}

\begin{figure*}[t]
    \centering
    \includegraphics[width=0.96\textwidth]{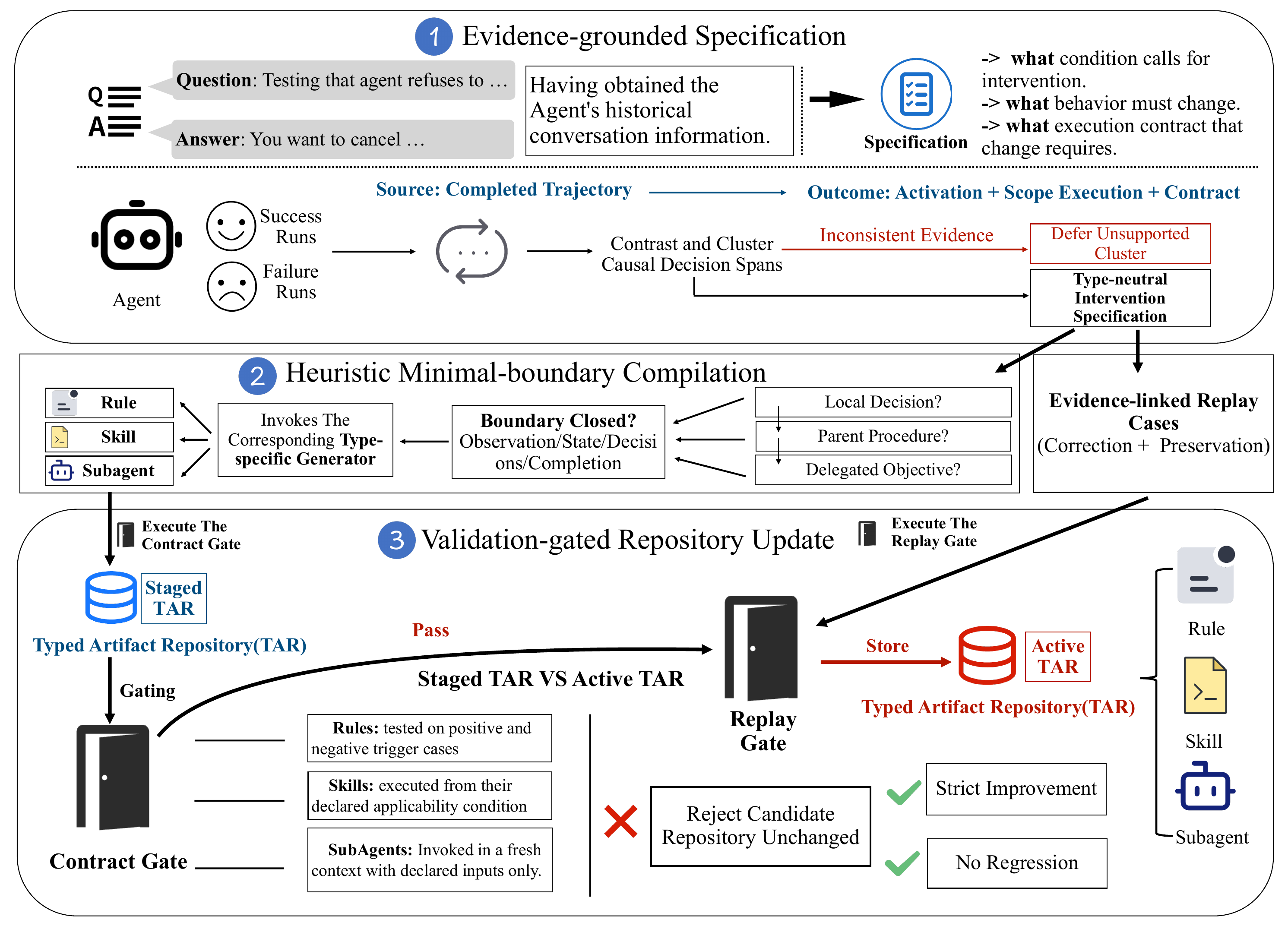}
    \caption{AutoRefine contrasts failed against successful trajectories to derive a type-neutral intervention specification, selects the first closed Rule, Skill, or Subagent in a runtime-relative ownership order, and admits the candidate only after contract and evidence-linked replay validation.}
    \label{fig:framework}
\end{figure*}

\subsection{Problem Setup}

An agent receives a sequence of tasks from a domain $\mathcal{D}$. Executing task $t_i$ produces a record $r_i=(\tau_i,f_i)$, where $\tau_i=\langle s_0,a_1,o_1,\ldots,a_T,o_T,s_T\rangle$ and terminal feedback $f_i\in\{0,1\}$. From a batch $\mathcal{B}=\{r_i\}_{i=1}^{n}$, AutoRefine constructs reusable external artifacts without updating the base model.

Let $\mathcal{H}$ be the space of observable histories and $\mathfrak{G}$ the space of corrective objectives. The analyzer maps an evidence-supported cluster to an intervention specification
\begin{equation}
\mathcal{I}=(\alpha,\beta,\mathcal{O},\mathcal{S},\mathcal{Q},\gamma,\mathcal{E}^{c},\mathcal{E}^{p}),
\label{eq:intervention_spec}
\end{equation}
where $\alpha,\gamma:\mathcal{H}\rightarrow\{0,1\}$ are activation and completion predicates; $\beta\in\mathfrak{G}$ is the corrective objective; $\mathcal{O}$ contains required observation fields; $\mathcal{S}$ pairs state variables with required lifetimes; $\mathcal{Q}$ contains dependent decisions; and $\mathcal{E}^{c},\mathcal{E}^{p}$ are sets of correction and preservation record--span identifiers supporting these fields. The specification is type-neutral: it states obligations without assigning their owner.

For artifact type $z\in\mathcal{Z}=\{\mathsf{Rule},\mathsf{Skill},\mathsf{Sub}\}$, a generated contract $C_z$ declares the observations, state, decisions, and completion semantics owned by that runtime object. We write
\begin{equation}
\operatorname{Close}(\mathcal{I},C_z)
=c_O\wedge c_S\wedge c_Q\wedge c_\gamma,
\label{eq:closure}
\end{equation}
where each $c$ is true only when the corresponding requirement in Eq.~\eqref{eq:intervention_spec} is both implemented by a declared contract field and grounded in a cited evidence span. A missing field or citation produces an insufficiency witness rather than a positive closure decision.

AutoRefine maintains a \emph{typed artifact repository} whose types are defined by runtime ownership rather than length or file format. A \textbf{Rule} owns one local decision by mapping an observable condition to a guard or prescription. A \textbf{Skill} owns a procedure: it organizes dependent decisions while remaining inside the parent agent's context and control loop. A \textbf{Subagent} owns a delegated objective: it is invoked through a declared input interface, executes a self-contained reasoning or tool-use loop with isolated working state, and terminates through a structured success or failure return. Artifact type is therefore an executable contract over state ownership, invocation, and standalone validation, not a label attached to generated text.

Figure~\ref{fig:framework} shows the resulting three-stage pipeline, which operates on batches of completed trajectories and not on the retrieval path of an individual test task. At deployment, Rules are injected into the parent context, while Skills and Subagents are installed as directory-scoped modules under the host runtime's native conventions; discovery and invocation routing are infrastructure rather than research contributions.

\subsection{Evidence-Grounded Intervention Specifications}

Single failures are poor compilation targets: their symptoms may be incidental, and a patch that explains one episode may damage behavior that already works. AutoRefine uses a trajectory batch as its evidence unit. For each failed record, an analyzer identifies the evaluator-visible failure, the earliest decision span that plausibly caused it, and the behavioral change required to avoid it. Records are grouped only when the same correction applies to their causal spans under the same observable conditions. Successful records from the same regime are retained as contrastive evidence that constrains the intervention with neighboring behavior to be preserved~\cite{fu2024autoguide,zhao2024expel}.

Each supported cluster instantiates Eq.~\eqref{eq:intervention_spec}. \emph{Activation} identifies the trigger and required observations; \emph{behavioral scope} contains the corrective objective and dependent decisions; and the \emph{execution obligation} identifies persistent state and observable completion.
Every field retains links to its supporting trajectory spans and to the associated correction and preservation tasks. These links exclude unsupported details and later instantiate replay cases. A cluster is deferred when the traces do not support one consistent trigger, correction, or completion condition. No artifact type is chosen at this stage, which prevents a generator from producing a role-shaped prompt first and retroactively naming it a Subagent.
The supplementary material gives the analyzer prompt structure, grouping criterion, and the division between language-model judgments and deterministic checks.

\subsection{Runtime-Relative First-Closure Compilation}

The selector is a heuristic \emph{compile-and-check} procedure rather than a learned classifier or an unconstrained request for a type label. For the evaluated runtime, it defines the search order $\mathsf{Rule}\mathrel{\triangleleft_{\!\rho}}\mathsf{Skill}\mathrel{\triangleleft_{\!\rho}}\mathsf{Sub}$: local decision, parent-executed procedure, then delegated objective. For each schema, a type-specific generator fills its contract slots using only evidence-linked fields, after which Eq.~\eqref{eq:closure} tests observation, state, decision, and completion closure. Selection is
\begin{equation}
z^{*}=\min_{\triangleleft_{\rho}}\{z:\operatorname{Close}(\mathcal{I},C_z)=1\},
\label{eq:first_closed}
\end{equation}
with deferral when the set is empty. A failed obligation becomes an \emph{insufficiency witness} carried into the next attempt.

Equation~\eqref{eq:first_closed} gives a conditional, auditable notion of minimality: $z^{*}$ is the first closed member of the declared runtime order for the extracted $\mathcal{I}$. It does not claim that the schemas are nested in every runtime or that the analyzer recovered every latent requirement. Because construction precedes checking, the selector cannot justify a Subagent by emitting a type label; it must produce the interface, state ownership, and termination semantics that make delegation executable.

Closure is a selection heuristic, not a verification oracle. The generator proposes contract slots and per-obligation citations, but the branch decision is mechanical: a schema closes only when every obligation resolves to a declared slot and to a citation that a deterministic controller matches against the input spans, so a bare type assertion cannot pass. Closure also never licenses a repository write on its own, and an artifact that certifies an unsuitable boundary must still survive execution-based contract tests and evaluator-scored replay.

\paragraph{Boundary attempts.}
The compiler first attempts a trigger--guard--response contract, which closes only when the trigger and guard are evaluable from parent-visible evidence, the response is localized to one control point, and its effect can be checked without retaining private intermediate state. It next attempts an ordered procedure, which closes when the parent already owns the required task state, can interleave the procedure with ongoing work, and remains responsible for deciding when the enclosing task is complete.

A Subagent is attempted only when the evidence defines a closed intermediate objective, sufficient declared inputs, and an observable return condition, while execution requires private state or a self-contained loop that should not be interleaved with the parent. Its contract declares invocation conditions, input and output schemas, isolated state and workspace, scoped tools, success and failure termination, and a structured return. Prompt length, a specialist persona, or placement in a separate system message satisfies none of these conditions; evidence of independent execution without a closed interface leads to deferral.

\begin{tcolorbox}[colback=gray!5,colframe=gray!50,title=\textbf{Worked Subagent: Lodging Package Planner},boxsep=2pt,left=4pt,right=4pt,top=2pt,bottom=2pt,before skip=5pt,after skip=5pt]
\footnotesize
\textbf{Evidence and invocation.} Repeated failures commit after a partial candidate scan, so once accommodation candidates exist the parent delegates the query constraints, assigned stays, party size, budget, and grounded candidate records.

\smallskip
\noindent\textbf{Owned execution and return.} In an isolated workspace, the artifact keeps a private candidate ledger, checks each listing against city, capacity, local nights, room type, house rules, and cost, and compares all surviving packages. It returns a selected package with its cost arithmetic or grounded exhaustion with exact blockers; a partial scan cannot terminate successfully.
\end{tcolorbox}
The Rule attempt fails because one local guard cannot own exhaustive package comparison. The Skill attempt leaves a state-lifetime and completion witness open: the ledger must stay isolated across a search that terminates only after full coverage. A budget pass/fail check has no independent comparison state and closes as a Rule even when implemented through a separate prompt.

After selection, the generator completes the chosen schema while preserving field-level links to supporting trajectories, and a traceability check rejects any required field that lacks evidence or depends on undeclared parent state. Selecting a different type therefore changes the executable unit and its validation surface, not a repository label; the supplementary material lists the required fields and complete examples for each schema.

\subsection{Validation-Gated Repository Updates}

Compilation writes a candidate to staging rather than to the active repository. Evidence links induce two compact replay sets drawn only from the learning partition: a \emph{correction set} of source failures and development cases with the same trigger and failure mechanism, and a \emph{preservation set} of contrastive successes and negative-boundary cases in which the trigger is observable but correct behavior should not change. Final evaluation cases never enter either set, so validation stays local to the claimed intervention.

Admission has two gates. The \emph{contract gate} tests whether the generated object implements its selected boundary: Rules are checked on positive and negative triggers; Skills are executed from their declared applicability condition to check that the declared inputs suffice and that the procedure reaches an observable result in the parent context; and Subagents are invoked in a fresh context using only declared inputs, with checks on I/O conformance, isolation from undeclared parent state, and explicit success and failure termination. Malformed artifacts are rejected before end-task evaluation.

The \emph{replay gate} compares the active repository $\mathcal{R}$ and staged repository $\mathcal{R}'$ on identical cases under the same evaluator and execution budget. Let $y_{\mathcal{R}}(x)\in[0,1]$ denote the evaluator outcome supplied by the configured replay protocol for case $x$ (binary for a single execution, or the protocol's aggregate when repeated outcomes are available). With correction and preservation sets $\mathcal{C}$ and $\mathcal{P}$, respectively, admission requires
\begin{equation}
\sum_{x\in\mathcal{C}}[y_{\mathcal{R}'}(x)-y_{\mathcal{R}}(x)]>0
\quad\land\quad
y_{\mathcal{R}'}(x)\ge y_{\mathcal{R}}(x)\;\;\forall x\in\mathcal{P}.
\label{eq:replay_gate}
\end{equation}
A tie on correction cases or any observed preservation regression rejects the candidate and leaves the active repository unchanged, so conflicting evidence resolves conservatively. The guarantee is deliberately case-bounded: it licenses the candidate on replayed evidence, not globally over $\mathcal{D}$.

An admitted artifact retains the replay cases induced by its source evidence. A revision is staged as a new version, tested on the predecessor's cases together with the evidence motivating the change, and activated only after both gates pass, which keeps regression scope explicit without full-repository replay. Pruning, similarity merging, retrieval, and global scheduling lie outside this mechanism.

\section{Experiments}

\subsection{Experimental Setup}

\paragraph{Evaluation design.} RQ1 measures end-to-end quality and online interaction cost, RQ2 isolates boundary availability, compilation, and gated admission, RQ3 asks whether successive repository writes accumulate capability, and RQ4 evaluates cross-domain reuse.

\paragraph{Benchmarks.}
RQ1--RQ3 use five task benchmarks: embodied execution (ALFWorld~\cite{shridhar2020alfworld}), scientific procedures (ScienceWorld~\cite{wang2022scienceworld}), constraint-heavy planning (TravelPlanner~\cite{xie2024travelplanner}), executable skills (SkillCraft~\cite{chen2026skillcraft}), and spreadsheet manipulation (SpreadsheetBench~\cite{ma2024spreadsheetbench}). They span local constraints, multi-step procedures, tool-mediated state changes, and decomposable objectives. RQ4 additionally uses the SkillFlow transfer suite~\cite{zhang2026skillflow}, whose five domains and topology-preserving task rewrites let us separate procedural reuse from lexical overlap.

\paragraph{Baselines.}
Baselines cover reasoning and reflection~\cite{yao2022react,shinn2023reflexion}, feedback-driven replanning~\cite{sun2023adaplanner}, trajectory learning~\cite{zhao2024expel,chen2024automanual,fu2024autoguide}, structured multi-agent execution~\cite{choi2025atlas}, and recent skill/subagent systems~\cite{yang2026skillopt,zhang2026agentfactory}. Within each benchmark, all methods share the GPT-5.6-terra backbone, evaluation split, evaluator, and execution budget, while retaining their method-specific prompts, memory, and artifact mechanisms.

\paragraph{Reporting convention.}
Our primary metric is task success under the benchmark evaluator. Table~\ref{tab:main_results} also reports mean assistant-message turns per case, counting top-level assistant responses and excluding tool results, offline construction, and AutoRefine's nested Subagent responses, which are audited separately. All entries are single-run point estimates under the fixed protocol. Ablations report percentage-point differences from Full AutoRefine, and component-level attribution comes from the controlled RQ2 interventions rather than from cross-method margins.

\paragraph{Protocol controls.}
Learning and evaluation data are disjoint, and repositories are frozen throughout held-out evaluation. The RQ1--RQ2 TravelPlanner studies use the same 180 evaluation queries (60 per difficulty) and an 80-iteration cap; artifact-availability conditions mask types from one 9-Rule/3-Skill/3-Subagent repository, while compiler ablations rebuild from the same learning evidence. Controlled conditions are compared as paired case-level outcomes with exact McNemar tests on discordant pairs. The supplementary material adds Wilson intervals, per-benchmark splits, leakage controls, and the paired SpreadsheetBench setup.

\subsection{RQ1: End-to-End Performance Across Domains}

\begin{table*}[!t]
\centering
\caption{End-to-end results: success (\%) / mean assistant-message turns per case, one decimal. Venue gives each baseline's original publication; the citation carries the year. Within each column, bold marks the best success and underline the second best when unique; tied best entries are all bold and turns are not ranked. The text reports count-derived values at two decimals.}
\label{tab:main_results}
\footnotesize
\setlength{\tabcolsep}{4.2pt}
\renewcommand{\arraystretch}{0.93}
\begin{tabular}{llccccc}
\toprule
Method & Venue & ALFWorld & ScienceWorld & TravelPlanner & SkillCraft & SpreadsheetBench \\
\midrule
ReAct~\cite{yao2022react} & ICLR & 79.9 / 14.0 & 71.5 / 26.2 & 34.4 / 26.1 & 83.3 / 9.8 & 41.1 / 6.4 \\
Reflexion~\cite{shinn2023reflexion} & NeurIPS & 85.1 / 15.7 & 72.2 / 42.0 & 34.4 / 77.9 & 83.3 / 13.2 & 44.3 / 9.1 \\
ReAct+Refl. & Combined & 88.1 / 16.1 & 72.9 / 40.2 & 37.2 / 80.2 & 88.9 / 14.0 & 48.2 / 9.8 \\
ExpeL~\cite{zhao2024expel} & AAAI & 84.3 / 16.0 & 74.3 / 21.5 & 45.0 / 18.0 & 86.1 / 8.7 & 45.0 / 6.8 \\
AdaPlanner~\cite{sun2023adaplanner} & NeurIPS & 82.1 / 14.6 & 77.1 / 18.8 & 44.4 / 14.9 & 86.1 / 7.9 & 47.5 / 6.1 \\
AutoManual~\cite{chen2024automanual} & NeurIPS & 96.3 / 13.5 & 78.5 / 19.0 & \underline{50.0} / 15.4 & 91.7 / 7.5 & 51.1 / 5.9 \\
AutoGuide~\cite{fu2024autoguide} & NeurIPS & 90.3 / 15.0 & 77.1 / 20.3 & 46.7 / 16.2 & 88.9 / 8.1 & 49.6 / 6.2 \\
ATLAS~\cite{choi2025atlas} & arXiv & 91.8 / 12.7 & 75.0 / 18.4 & 48.3 / 9.5 & 88.9 / 6.9 & 52.1 / 5.4 \\
SkillOpt~\cite{yang2026skillopt} & arXiv & \underline{97.8} / 18.6 & \underline{79.9} / 17.2 & 47.2 / 12.4 & \textbf{100.0} / 2.1 & \underline{67.5} / 4.0 \\
AgentFactory~\cite{zhang2026agentfactory} & arXiv & 95.5 / 14.8 & 76.4 / 16.0 & 43.3 / 6.2 & \textbf{100.0} / 3.0 & 59.6 / 5.1 \\
\midrule
\textbf{AutoRefine} & This work & \textbf{100.0} / 11.9 & \textbf{80.6} / 16.5 & \textbf{80.6} / 20.7 & \textbf{100.0} / 2.4 & \textbf{73.6} / 3.1 \\
\bottomrule
\end{tabular}
\end{table*}

RQ1 asks whether the complete system improves held-out execution and whether the measured pattern is confined to a single benchmark. Table~\ref{tab:main_results} reports success together with the online interaction cost under the unified evaluation harness.

\paragraph{Results.} AutoRefine records the highest success on ALFWorld (134/134), ScienceWorld (116/144), TravelPlanner (80.56\%, 145/180), and SpreadsheetBench (73.57\%, 206/280), and ties the SkillCraft ceiling at 36/36. Against the strongest-success baseline it also uses fewer top-level turns on ALFWorld, ScienceWorld, and SpreadsheetBench, and 0.3 turns more on the saturated SkillCraft. TravelPlanner is the only setting where the 30.6-point gain comes with more turns (20.7 vs. 15.4).

\paragraph{Analysis.} The margin tracks the room a benchmark leaves. ALFWorld and SkillCraft sit within a few points of their ceilings and ScienceWorld separates the top two methods by 0.7 points, whereas every baseline on TravelPlanner stays between 34.4\% and 50.0\%. A TravelPlanner query counts as solved only when one itinerary satisfies every commonsense requirement and hard constraint at once, so a single violation discards an otherwise complete plan, which prompt-level advice rarely prevents. The per-difficulty counts in the supplementary material locate the 30.6-point gain in that structure: Skills carry the plan-construction routine in every bucket, while Rules become decisive only on Hard queries, where several constraints interact. We read this margin as evidence about unsaturated multi-constraint planning, not as a uniform expected gain, and the turn results support no comparable claim about cost. Of the 20.68 parent turns per case, 1.48 invoke a Subagent, whose 8.79 internal responses the main table excludes, giving an expanded actor count of 29.47. Disabling Subagents raises parent turns to 34.03 while success drops to 72.22\%. The supplementary material reports the full call, token, and wall-time audit.

\subsection{RQ2: Which Design Choices Make Typed Compilation Effective?}

End-to-end gains may follow from exposing the agent to more text rather than selecting a representation. RQ2 therefore intervenes on boundary availability, compilation, and admission.

\paragraph{Ablation protocol.} All conditions in Table~\ref{tab:representation_control} share the TravelPlanner protocol. \emph{Artifact availability} masks types from a frozen repository; \emph{heuristic compilation} rebuilds it after replacing intervention-first selection, minimality, or closure; and \emph{validation} keeps candidates fixed but removes a gate or preservation cases. The three groups therefore separate artifact utility from construction and admission policy. Paired tests cover the type-removal conditions, for which per-case ledgers are retained; the No-Artifact control is an aggregate-only record, so we report its rate but no paired test.

\begin{table}[t]
\centering
\caption{TravelPlanner ablations under the shared evaluation protocol. Avg.\ turns are mean top-level parent responses per case: measured counts for the three type-removal rows, an aggregate-only record for No Artifact, and ledger-derived means elsewhere, reported for reference only. $\Delta$ Pass comes from case counts, except for No Artifact, whose delta is a difference of rates. The appendix adds per-difficulty counts and delegation costs.}
\label{tab:representation_control}
\scriptsize
\setlength{\tabcolsep}{3.5pt}
\begin{tabular}{llrrr}
\toprule
Group & Condition & Pass (\%) & Avg. turns & $\Delta$ Pass \\
\midrule
Reference & \textbf{Full AutoRefine} & \textbf{80.56} & 20.68 & -- \\
\midrule
\multirow{4}{*}{\shortstack[l]{Artifact\\availability}}
& w/o Rule & 79.44 & 30.71 & -1.11 \\
& w/o Skill & 60.56 & 10.78 & -20.00 \\
& w/o Subagent & 72.22 & 34.03 & -8.33 \\
& No Artifact & 34.44 & 21.07 & -46.12 \\
\midrule
\multirow{3}{*}{\shortstack[l]{Heuristic\\compilation}}
& Direct Type Selection & 74.44 & 20.15 & -6.11 \\
& w/o Minimality Order & 69.44 & 21.74 & -11.11 \\
& w/o Boundary-Closure Check & 65.56 & 19.01 & -15.00 \\
\midrule
\multirow{3}{*}{Validation}
& w/o Contract Gate & 75.00 & 20.42 & -5.56 \\
& w/o Replay Gate & 64.44 & 18.30 & -16.11 \\
& w/o Preservation Cases & 70.00 & 19.54 & -10.56 \\
\bottomrule
\end{tabular}
\end{table}

\paragraph{Artifact complementarity.} Removing all artifacts reduces success from 80.56\% to 34.44\%. Turns track prescribed checking rather than task difficulty: with no artifacts the agent also skips artifact-prescribed verification, spending fewer turns (21.07) than Rule removal alone (30.71) while succeeding far less often. Skills contribute 20.00 points (paired exact McNemar $p<.001$) and Subagents 8.33 ($p=.040$), the latter concentrated on Medium and Hard tasks, consistent with delegation helping longer plans. Removing Rules changes aggregate success by only 1.11 points ($p=.875$) but loses 6.67 points on Hard tasks, so Rules act as targeted guards rather than a uniformly dominant representation. The supplementary material reports the per-bucket counts.

\paragraph{Compilation and admission.} Direct type selection loses 6.11 points, removing minimality loses 11.11, and omitting closure loses 15.00: offering three formats is insufficient unless the selected owner closes over the required execution contract. Removing the contract gate, replay gate, or preservation cases costs 5.56, 16.11, and 10.56 points. The large replay loss shows that interface validity cannot replace behavioral evidence; a locally valid candidate may still disrupt solved tasks.

\subsection{RQ3: Does Gated Admission Prevent Repository Degradation?}

Final quality cannot reveal whether intermediate updates accumulated capability or merely recovered after destructive writes. This study therefore runs on TravelPlanner, using its 180 annotated validation queries as one common ordered learning stream for all methods; the 45 official training queries of RQ1--RQ2 are too few to expose repeated writes, and the validation queries carry the same annotations and structure. Checkpoints every 20 tasks are scored on a fixed held-out set drawn from the disjoint test split, so the queries that carry the learning signal are never scored and no repository built here enters the RQ1--RQ2 evaluations. We report held-out pass rate, repository size, artifact use rate, and ledger-derived signed degradation, where positive values indicate net loss within the baseline-success ledger and negative values net recovery rather than absence of individual regressions. AgentFactory tests horizontal proliferation, SkillOpt vertical overwrite of one Skill, and AutoRefine heterogeneous coexistence with evidence-linked replay. The retained checkpoint records are aggregates without case identifiers, so this stream diagnoses accumulation within the stream and supports no paired test.

\begin{figure}[H]
\centering
\includegraphics[width=\columnwidth]{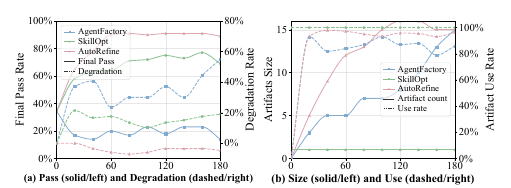}
\caption{Repository dynamics on the common TravelPlanner learning stream. (a) Final held-out pass rate (solid, left axis) and ledger-reported signed degradation (dashed, right axis), where negative AutoRefine values denote net recovery. (b) Repository size (solid, left axis) and artifact use rate (dashed, right axis).}
\label{fig:repository_evolution}
\end{figure}

\paragraph{Longitudinal results.} Figure~\ref{fig:repository_evolution} traces the three regimes. From the common 34.44\% start, AgentFactory falls to 17\% after its first three artifacts and ends at 14\% with fifteen stored artifacts, while signed degradation rises to 55.0\%; its artifacts are invoked in 79--93\% of evaluations throughout, so high use measures exposure and not utility. SkillOpt reaches 77\%, but its final rewrite of the single maintained Skill reduces performance to 69\% as signed degradation returns to 19.0\%. AutoRefine reaches 65\% after 20 learning tasks and 89--91\% from 60 onward, with 12 to 16 artifacts and 93.2--97.2\% use, and its net signed degradation stays non-positive (0 to $-7.1\%$). Growth alone does not explain the difference: AgentFactory and AutoRefine end with comparable artifact counts but diverge in whether the writes survive as held-out capability.

\paragraph{Admission quality and cost.} On SpreadsheetBench, gated updates reach 73.57\% against 67.50\% for the SkillOpt control under a shared candidate stream and split (paired exact McNemar $p=.027$), while offline consumption rises from 3.34M to 9.43M tokens. The margin compares accumulation regimes; Table~\ref{tab:representation_control} isolates the gates. Validation therefore shifts computation to repository construction; we claim more reliable accumulation, not universally cheaper learning. The supplementary material gives the cost breakdown.

\subsection{RQ4: Do Artifacts Generalize Across Domains?}

Frozen source repositories are evaluated on SkillFlow's Domain-Agnostic Execution Flow without target-side updates, across Finance \& Economics (FE), Operations \& Supply Chain (OS), Healthcare \& Life Sciences (HLS), Governance \& Strategy (GS), and Data \& Document Intelligence (DDI), yielding a $5\!\times\!5$ matched- and cross-domain matrix.

\begin{table}[t]
\centering
\caption{SkillFlow transfer: absolute target gains (percentage points) from each frozen source repository over the target-only baselines in the last row. Bold entries are matched-domain transfer.}
\label{tab:transfer_matrix}
\scriptsize
\setlength{\tabcolsep}{5pt}
\begin{tabular}{lrrrrr}
\toprule
Source $\backslash$ Target & FE & OS & HLS & GS & DDI \\
\midrule
FE  & \textbf{15.8} & 8.1 & 9.7 & 5.4 & 7.6 \\
OS  & 6.3 & \textbf{13.9} & 6.8 & 3.1 & 8.2 \\
HLS & 7.9 & 1.4 & \textbf{15.6} & 7.8 & 10.1 \\
GS  & 4.5 & 2.7 & 11.2 & \textbf{16.0} & 8.4 \\
DDI & 8.6 & 7.5 & 8.3 & 6.2 & \textbf{9.7} \\
\midrule
Target-only baseline & 52.4 & 72.6 & 72.3 & 80.4 & 78.4 \\
\bottomrule
\end{tabular}
\end{table}

\paragraph{Transfer results.} All 25 source--target pairs in Table~\ref{tab:transfer_matrix} improve, by 1.4--16.0 points (8.43 on average). Matched-domain gains average 14.20 points against 6.99 off diagonal, and the matched repository is the strongest source for four of the five targets. The exception is DDI, where HLS transfers slightly better (10.1 against 9.7), so domain-specific evidence helps on average without dominating every target, while the uniformly positive off-diagonal gains indicate that workflow structure itself transfers. Per-target and per-source means appear in the supplementary material.

\subsection{Error Analysis}

Results expose two error modes with different remedies. A \emph{boundary error} assigns an intervention to an owner that cannot observe, retain, control, or complete it; the lodging example is the canonical case, where a parent-executed procedure loses the isolated ledger that its completion test requires. First-closure compilation targets this mode, which is why closure is the costliest compilation intervention. An \emph{evidence-scope error} activates a structurally valid artifact outside the cases that licensed it, and the replay ablation, the unchecked degradation in RQ3, and the off-diagonal transfer gap each expose it. Preservation-aware replay bounds this mode at admission, but only over cases linked to the artifact's own evidence.

\FloatBarrier
\section{Conclusion}

This work reframes trajectory learning as the construction of typed, testable agent capabilities: AutoRefine derives an evidence-linked intervention specification from contrastive trajectories, compiles it into the first closed Rule, Skill, or Subagent under a runtime-relative ownership order, and admits it through contract and replay gates. Artifact type fixes what an intervention observes, what it owns, and how it can be tested.

Across five benchmarks, typed repositories improve or match end-task performance, with the largest gain where local constraints, reusable procedures, and bounded delegation coexist. Neither content nor validation alone suffices: removing boundary closure or replay causes the largest losses among the policies we ablate. Longitudinal checkpoints separate repository activity from improvement, and transfer is positive on all 25 pairs while favoring matched-domain evidence.

These findings suggest a stricter standard for self-evolving agents: trajectory evidence motivates a candidate capability, an execution contract states what it owns, and behavioral validation decides whether it may be activated. Three limits remain: closure is only as complete as the extracted specification, replay licenses an artifact on its linked cases rather than the task distribution, and both gates judge one artifact at a time, so artifact interactions need joint replay.

\begingroup
\small
\bibliography{ref}
\endgroup

\clearpage
\onecolumn
\appendix

\section{Extended Methodological Details}\label{app:additional_discussions}

This appendix provides the operational details needed to interpret and reproduce AutoRefine. We first distinguish typed artifact compilation from fixed-representation experience learning, then specify the contracts used by the compiler and validation gates. We next report the frozen evaluation protocols and supporting analyses, and conclude with limitations and broader impacts.

\subsection{Relation to Prior Experience-Learning Systems}\label{app:diff_prior}

AutoRefine differs from prior work in the decision made \emph{before} an external capability is written. Experience-memory methods typically choose an insight, guideline, or manual as the storage unit~\cite{zhao2024expel,fu2024autoguide,chen2024automanual}; skill optimizers improve a single procedural representation~\cite{yang2026skillopt,wang2026skillx,lin2026museautosk}; and AgentFactory preserves solutions as executable Subagents~\cite{zhang2026agentfactory}. Auto-harness systems expose a wider set of editable files and validate the resulting harness~\cite{zhang2026selfharness,lin2026ahe}. These systems provide important extraction, optimization, and validation mechanisms, but the representation is either fixed in advance or selected as an unconstrained harness-edit target. AutoRefine instead constructs an artifact-type-neutral intervention and asks which execution boundary is sufficient to own it. Table~\ref{tab:appendix_positioning} contrasts the construction targets and validation obligations.

\begin{table}[H]
\centering
\caption{Positioning by construction target and validation obligation. ``Fixed'' means that the reusable unit is selected before the evidence for an individual intervention is analyzed.}
\label{tab:appendix_positioning}
\small
\setlength{\tabcolsep}{5pt}
\begin{tabularx}{0.97\textwidth}{
>{\raggedright\arraybackslash}p{0.20\textwidth}
>{\raggedright\arraybackslash}p{0.20\textwidth}
>{\raggedright\arraybackslash}p{0.20\textwidth}
>{\raggedright\arraybackslash}X}
\toprule
Approach family & Primary reusable unit & Representation decision & Validation surface \\
\midrule
Experience memory~\cite{zhao2024expel,fu2024autoguide,chen2024automanual}
& Insight, guideline, or manual
& Fixed textual form
& Method-specific task feedback \\
Skill optimization~\cite{yang2026skillopt,wang2026skillx,lin2026museautosk}
& Skill or skill package
& Fixed procedural form
& Skill execution and held-out utility \\
Executable Subagents~\cite{zhang2026agentfactory}
& Isolated program or Subagent
& Fixed delegated form
& Subagent execution, often on the inducing task \\
Auto-harness editing~\cite{zhang2026selfharness,lin2026ahe}
& Heterogeneous harness files
& Edit target proposed within the harness
& Whole-harness prediction or regression tests \\
\textbf{AutoRefine}
& Rule, Skill, or bounded Subagent
& First closed boundary in a runtime-relative ownership order
& Type contract plus correction and preservation replay \\
\bottomrule
\end{tabularx}
\end{table}

This distinction is deliberately narrower than a claim that prior systems lack validation. AutoRefine's contribution is the coupling between representation and validation: selecting a type determines which observations and state the artifact owns, which decisions it may make, how it terminates, and which standalone tests are meaningful. Repository search, directory discovery, and invocation routing do not determine this boundary and are outside the studied mechanism.

\subsection{Formal Objects and Runtime-Relative Boundary Order}

Let $\mathcal{H}$ be the space of observable execution histories, $\mathfrak{G}$ the space of corrective objectives, and $\mathfrak{O}$, $\mathfrak{V}$, $\mathfrak{L}$, and $\mathfrak{Q}$ the universes of observation fields, state variables, lifetimes, and decisions. An intervention specification is the typed object
\begin{equation}
\mathcal{I}=(\alpha,\beta,\mathcal{O},\mathcal{S},\mathcal{Q},\gamma,\mathcal{E}^{c},\mathcal{E}^{p}),
\label{eq:app_intervention_spec}
\end{equation}
where $\alpha,\gamma:\mathcal{H}\rightarrow\{0,1\}$ are activation and completion predicates, $\beta\in\mathfrak{G}$ is the corrective objective, $\mathcal{O}\subseteq\mathfrak{O}$ lists required observations, $\mathcal{S}\subseteq\mathfrak{V}\times\mathfrak{L}$ pairs state variables with required lifetimes, $\mathcal{Q}\subseteq\mathfrak{Q}$ lists dependent decisions, and $\mathcal{E}^{c},\mathcal{E}^{p}$ are sets of record--span identifiers linking these fields to correction and preservation evidence. The compiler may instantiate an artifact only from fields supported by these links.

For schema $z\in\mathcal{Z}=\{\mathsf{Rule},\mathsf{Skill},\mathsf{Sub}\}$, contract $C_z$ declares its observable inputs, owned state, controlled decisions, completion test, and schema-specific interface. Closure is the conjunction
\begin{equation}
\operatorname{Close}(\mathcal{I},C_z)=c_O\wedge c_S\wedge c_Q\wedge c_\gamma,
\label{eq:app_closure}
\end{equation}
where an obligation is true only when the contract implements it and every required field resolves to cited evidence. This definition separates two failure modes: a missing runtime capability produces an ownership witness, whereas a missing or invalid citation produces an evidence witness.

The evaluated runtime imposes the search order $\mathsf{Rule}\mathrel{\triangleleft_{\!\rho}}\mathsf{Skill}\mathrel{\triangleleft_{\!\rho}}\mathsf{Sub}$, ranking a local parent decision, a parent-executed procedure, and a delegated objective by additional ownership. The selected type is
\begin{equation}
z^*=\min_{\triangleleft_\rho}\{z:\operatorname{Close}(\mathcal{I},C_z)=1\};
\label{eq:app_selection}
\end{equation}
if the set is empty, the intervention is deferred. First closure is thus minimal only relative to the extracted $\mathcal{I}$ and the declared runtime order. The three schemas are not asserted to form a universal expressivity lattice: a runtime that cannot pass required observations, preserve parent interleaving, or enforce Subagent isolation may make schemas incomparable, in which case its order and closure checks must be adapted.

\subsection{Typed Contracts and Boundary Closure}

Table~\ref{tab:artifact_contracts} expands the three schemas used by the compiler. The schemas are not alternative serialization formats. Each introduces a different owner and consequently a different failure surface. A Rule cannot hide state or make a sequence of dependent choices; a Skill may coordinate such choices but remains inside the parent's context and control loop; a Subagent receives a closed objective through a declared invocation boundary and returns after running an isolated loop.

\begin{table}[H]
\centering
\caption{Required fields and standalone checks for the three artifact contracts.}
\label{tab:artifact_contracts}
\small
\setlength{\tabcolsep}{5pt}
\begin{tabularx}{0.97\textwidth}{
>{\raggedright\arraybackslash}p{0.08\textwidth}
>{\raggedright\arraybackslash}p{0.21\textwidth}
>{\raggedright\arraybackslash}p{0.32\textwidth}
>{\raggedright\arraybackslash}X}
\toprule
Type & Runtime ownership & Required contract fields & Contract-gate checks \\
\midrule
Rule
& One parent-visible decision
& Trigger, guard or prescription, response, local check
& Positive and negative triggers; response is locally decidable; no hidden state \\
Skill
& Parent-executed procedure
& Applicability, inputs, ordered steps, expected result, recovery
& Declared inputs suffice; steps execute in the parent context; result is observable \\
Subagent
& Delegated intermediate objective
& Invocation, I/O schemas, private state, scoped tools, success and failure termination, structured return
& Fresh-context invocation; state isolation; I/O conformance; explicit termination; no undeclared parent dependency \\
\bottomrule
\end{tabularx}
\end{table}

For every attempted schema, boundary closure evaluates the four obligations in Eq.~\eqref{eq:app_closure}. \emph{Observation closure} asks whether the proposed owner can observe every required input. \emph{State closure} asks whether it can retain each variable for the lifetime implied by the intervention. \emph{Decision closure} asks whether all dependent decisions fall within its control. \emph{Completion closure} asks whether it can determine and report when the intervention is complete. A failed obligation is recorded as an insufficiency witness rather than collapsed into a scalar score. The next schema in $\triangleleft_\rho$ must close that witness in addition to the remaining obligations. The first closed construction is therefore minimal in the precise sense of Eq.~\eqref{eq:app_selection}, not necessarily the shortest generated text or the least expressive object in another runtime.

\subsection{Evidence-to-Boundary Example}

The lodging example in the main paper illustrates how the selector reasons about ownership. Several TravelPlanner failures committed to a lodging package after inspecting only part of the returned candidate set. The corresponding successes retained a candidate-level ledger, checked each listing against the assigned city, party capacity, local stay length, room type, explicit house rules, and total cost, and terminated only after every candidate had been accounted for.

The artifact-type-neutral intervention specification records: (i) activation after grounded accommodation candidates exist and before a package is committed; (ii) the corrective objective of exhaustive compliant-package comparison; and (iii) a ledger whose lifetime extends from the first candidate check to a selected package or grounded exhaustion. The Rule attempt fails decision and state closure because a local guard such as ``check all candidates'' neither owns the ledger nor performs the comparison. The Skill attempt can describe the checks, but leaves their private intermediate state and independent completion inside the parent loop. The Subagent attempt closes both witnesses by receiving a complete candidate set, owning a private ledger, and returning only after full coverage. This conclusion follows from the required execution boundary, not from the length or tone of the generated instruction.

\subsection{Illustrative Artifact Contracts}\label{app:subagent_examples}

The following examples show the runtime fields produced after type selection. They are presented to clarify the contract, not as additional performance claims.

\noindent\textbf{Rule: stop after confirmed delivery.}\par
\begin{tcolorbox}[colback=green!3,colframe=green!35,boxsep=2pt,left=4pt,right=4pt,top=3pt,bottom=3pt]
\small
\textbf{Activation:} the current observation identifies the required object, required state, and target receptacle. \textbf{Guard:} the observation explicitly confirms that the state-modified object is in or on the requested receptacle. \textbf{Response:} terminate the episode without further manipulation. \textbf{Local check:} fire on confirmed placement; do not fire when the object, state, or receptacle differs. No private state or delegated objective is introduced.
\end{tcolorbox}

\noindent\textbf{Skill: heat then place an object.}\par
\begin{tcolorbox}[colback=blue!3,colframe=blue!35,boxsep=2pt,left=4pt,right=4pt,top=3pt,bottom=3pt]
\small
\textbf{Applicability and inputs:} an embodied task names an object, the required heated state, and a destination; the parent exposes navigation and manipulation actions. \textbf{Procedure:} search receptacles systematically, acquire the grounded object, navigate to a compatible appliance, heat and verify it, deliver it to the requested receptacle, and confirm placement. \textbf{Recovery:} after an invalid action, refresh the local observation and resume from the last verified stage rather than restarting blindly. \textbf{Expected result:} the parent observes successful placement and remains responsible for task termination. The procedure spans dependent choices, but neither isolates state nor owns an independent objective.
\end{tcolorbox}

\paragraph{Subagent: exhaustive lodging package planner.}
The contract below is based on a validated TravelPlanner artifact and includes every field that separates delegation from prompt augmentation.

\begin{tcolorbox}[colback=orange!3,colframe=orange!45,title=\textbf{Exhaustive Lodging Package Planner},breakable,boxsep=2pt,left=4pt,right=4pt,top=3pt,bottom=3pt]
\small
\textbf{Invocation boundary.} Invoke after accommodation candidates have been grounded for one or more assigned overnight stays and before the parent commits to a lodging package. Skip invocation when no overnight stay is required or the exact package has already been validated.

\textbf{Input schema.}
\begin{itemize}[leftmargin=1.5em,topsep=2pt,itemsep=1pt]
    \item complete query constraints, dates, party size, and total budget;
    \item assigned city stays and the consecutive nights in each city;
    \item complete grounded accommodation records, including listing identity, structured city, room type, nightly price, explicit house rules, local minimum nights, and maximum occupancy;
    \item relevant parent evidence records, but no unlisted parent reasoning state.
\end{itemize}

\textbf{Owned state and execution.} The Subagent creates a private candidate ledger. For each listing, it records evidence, computes the required number of units as $\lceil\text{party size}/\text{maximum occupancy}\rceil$, checks the assigned local stay rather than the full trip against minimum nights, applies only explicit matching house-rule prohibitions, and computes the package cost. It must compare every surviving package and may not terminate after finding the first valid option. Candidate-level calculations remain isolated until the final return.

\textbf{Scoped evidence and tools.} The Subagent may read only the supplied accommodation records and query fields, and may write candidate-level checks to its local ledger. It cannot invent listings, infer a city from a listing name when a structured city field exists, or use undeclared parent memory.

\textbf{Success termination.} Every candidate is classified, at least one complete package satisfies the declared constraints, and the lowest-cost compliant package is selected. The return includes the selected listing or listings, unit and night arithmetic, total cost, budget impact, and the rejected alternatives with exact reasons.

\textbf{Failure termination.} If no package satisfies all constraints, the Subagent returns \texttt{exhausted} rather than a fabricated selection. The return includes full candidate coverage, exact blockers, the least-conflicting grounded alternative when one exists, budget impact, and unresolved items. Missing required input produces \texttt{blocked} with the missing fields.

\textbf{Structured return schema.}
\begin{lstlisting}[numbers=none,xleftmargin=1em,basicstyle=\footnotesize\ttfamily]
status: success | exhausted | blocked
selected_package: [listing_id, city, units, nights, total_cost] | null
candidate_audit: [{listing_id, checks, arithmetic, decision}]
budget_impact: {package_cost, remaining_budget}
blockers: [grounded constraint violations]
unresolved_items: [missing or ambiguous evidence]
\end{lstlisting}

\textbf{Standalone validation.} Invoke the artifact in a fresh context using only the declared inputs; verify that every candidate appears exactly once in the audit, recompute capacity and cost arithmetic, deny access to undeclared parent state, and test both a feasible package and grounded exhaustion. A partial scan cannot return \texttt{success}.
\end{tcolorbox}

A budget threshold check or output-format check does not satisfy this contract. Even if implemented through a separate system message, it observes a fixed input, returns a local pass/fail decision, and owns neither a private comparison state nor an independently terminating objective. AutoRefine therefore compiles such a check as a Rule. This nearby counterexample prevents implementation details such as prompt placement from defining the artifact type.

\subsection{Validation-Gated Admission and Revision}

Compilation never writes directly to the active repository. Each candidate first enters staging with links to its inducing failure spans, contrastive successful spans, and the tasks that instantiate its claimed activation boundary. These links induce a correction set and a preservation set. Correction cases exercise the failure mechanism the artifact claims to repair; preservation cases contain neighboring successes and negative-boundary cases in which the artifact must not change already-correct behavior.

The contract gate evaluates the structural obligations in Table~\ref{tab:artifact_contracts}. The replay gate then runs the active repository $\mathcal{R}$ and staged repository $\mathcal{R}'$ on identical cases under the same execution budget and evaluator. Let $y_{\mathcal{R}}(x)\in[0,1]$ be the evaluator outcome returned by the configured replay protocol and let $d(x)=y_{\mathcal{R}'}(x)-y_{\mathcal{R}}(x)$. The admission decision is
\begin{equation}
\operatorname{Admit}(\mathcal{R}',\mathcal{R})=\operatorname{Contract}(C_z)
\wedge\sum_{x\in\mathcal{C}}d(x)>0
\wedge\bigwedge_{x\in\mathcal{P}}[d(x)\ge0].
\label{eq:app_admission}
\end{equation}
Here $\mathcal{C}$ and $\mathcal{P}$ are the correction and preservation sets. For a single execution $y$ is binary; if a configured protocol supplies an aggregate over repeated outcomes, the same rule operates on that aggregate. The decision rule therefore does not require an assumed repeat count. A correction tie, a malformed contract, or any observed preservation regression rejects the candidate; preservation takes precedence when the two objectives conflict. The guarantee is intentionally evidence-bounded: passing replay licenses the candidate on the associated cases, not on every task in the domain.

Revisions follow the same path. A proposed revision is tested on the predecessor's associated cases together with the new evidence that motivated the change, and only the staged version becomes active after both gates pass. This mechanism provides artifact-local regression obligations without metadata-based pruning, similarity merging, or a global task--artifact graph. Those repository-engineering policies are not required by the method evaluated in this paper. Algorithm~\ref{alg:autorefine} states the full compilation and admission procedure.

\begin{algorithm}[H]
\caption{Evidence-linked typed compilation and admission}
\label{alg:autorefine}
\begin{algorithmic}[1]
\REQUIRE trajectory batch $\mathcal{B}$, active repository $\mathcal{R}$, runtime order $\triangleleft_\rho$
\ENSURE updated repository $\mathcal{R}$
\STATE $Q\leftarrow\textsc{AnalyzeAndNormalize}(\mathcal{B})$
\STATE $\mathcal{G}\leftarrow\textsc{CompatibleClusters}(Q)$
\FOR{cluster $G\in\mathcal{G}$}
  \STATE $\mathcal{I}\leftarrow\textsc{BuildSpecification}(G)$
  \STATE $a\leftarrow\varnothing$
  \FOR{$z$ in $\mathsf{Rule}\triangleleft_\rho\mathsf{Skill}\triangleleft_\rho\mathsf{Sub}$}
    \STATE $(a,C_z,w)\leftarrow\textsc{GenerateAndCheck}(\mathcal{I},z)$
    \IF{$\operatorname{Close}(\mathcal{I},C_z)$}
      \STATE \textbf{break}
    \ENDIF
  \ENDFOR
  \IF{$a\neq\varnothing$ and $\operatorname{Admit}(\mathcal{R}\oplus a,\mathcal{R})$}
    \STATE $\mathcal{R}\leftarrow\mathcal{R}\oplus a$
  \ENDIF
\ENDFOR
\RETURN $\mathcal{R}$
\end{algorithmic}
\end{algorithm}

\section{Experimental Protocols and Additional Analyses}\label{app:experimental_setup}

\subsection{Benchmarks and Data Separation}\label{app:dataset_details}

All held-out evaluations use frozen repositories: evaluation trajectories never produce, revise, or admit artifacts. Table~\ref{tab:benchmark_protocols} summarizes the data partitions and the controls most relevant to leakage and comparability.

\begin{table}[H]
\centering
\caption{Benchmark-specific protocols. A dash indicates that the released benchmark uses task families or variations rather than a single shared item count.}
\label{tab:benchmark_protocols}
\small
\setlength{\tabcolsep}{5pt}
\begin{tabularx}{0.98\textwidth}{
>{\raggedright\arraybackslash}p{0.16\textwidth}
>{\raggedright\arraybackslash}p{0.14\textwidth}
>{\raggedright\arraybackslash}p{0.15\textwidth}
>{\raggedright\arraybackslash}p{0.24\textwidth}
>{\raggedright\arraybackslash}X}
\toprule
Benchmark & Learn / dev & Held-out evaluation & Separation control & Execution budget and evaluator \\
\midrule
ALFWorld~\cite{shridhar2020alfworld}
& 39 / -- episodes & 134 \texttt{valid\_unseen}
& No \texttt{valid\_unseen} write; repository frozen before evaluation
& 50 environment steps; benchmark success signal \\
ScienceWorld~\cite{wang2022scienceworld}
& Official train variations & 144 held-out test variations
& Selected Boil and Thermometer families; no evaluation-time writes
& 50 actions; environment goal completion \\
TravelPlanner~\cite{xie2024travelplanner}
& RQ1--RQ2: 45 official training queries. RQ3: the 180 annotated validation queries
& RQ1--RQ2: 180 validation queries. RQ3: held-out cases from the disjoint test split
& Fixed 60 Easy / 60 Medium / 60 Hard cases for all controlled conditions; RQ3 repositories are never reused in RQ1--RQ2
& 80 iterations; official offline commonsense and hard-constraint checks \\
SkillCraft
& 11 / 4 families & 6 families (36 instances)
& Split by base task family, not templated instance
& Executed-output verifier; repository frozen \\
SpreadsheetBench
& 80 / 40 cases & 280 cases
& Disjoint item IDs, workbook paths, and workbook pairs
& Three agent turns; verified target-cell and sheet checks \\
\bottomrule
\end{tabularx}
\end{table}

\paragraph{ALFWorld and ScienceWorld.}
ALFWorld evaluates six household task families involving object search, state modification, and placement. The repository-learning sample contains 39 training episodes, of which 37 succeed; evaluation covers all 134 \texttt{valid\_unseen} tasks. ScienceWorld evaluates procedural scientific reasoning in the Boil and Thermometer families, which require instrument search, state tracking, and observation-dependent action sequences. We keep the benchmark's own variation split, so learning uses training variations and evaluation covers the 144 held-out test variations of the two families (9 for \texttt{boil} and 135 for \texttt{use-thermometer}). For both environments, success is determined by the simulator rather than a language-model judge.

\paragraph{TravelPlanner.}
TravelPlanner requires a grounded multi-day itinerary satisfying transportation, accommodation, dining, attraction, budget, and local constraints. The RQ2 artifact-availability study freezes one repository containing 9 Rules, 3 Skills, and 3 Subagents; compiler ablations instead rebuild from the same learning evidence under the stated policy change. All RQ1--RQ2 variants use the same 180 evaluation queries, SingleAgent runner, GPT-5.6-terra backbone, 80-iteration cap, and offline official evaluator. Planner delivery validation is disabled, and evaluator feedback is never returned to the executing agent. The dataset and full-repository SHA-256 values are \texttt{4b36f2f0301e3e6bd6aea2dc7d919b9f434f59d05618256a7fba5955c76dffc1} and \texttt{159424f3286c77ff945d7248e00408be59ac2db61653e55a35df5e1d683d4145}, respectively.

The longitudinal study (RQ3) uses a different partition of the same benchmark, because 45 training queries cannot support a 180-task write stream. Its ordered learning stream is the 180 annotated validation queries, which carry the same reference annotations and query structure as the official training split, and its checkpoints are scored on held-out cases from the disjoint test split. The two partitions are therefore never mixed within one experiment: repositories built from the RQ3 stream are used only for RQ3 checkpoints, and the RQ1--RQ2 repositories, built from the 45 training queries, are the only ones evaluated on the 180 validation queries. The retained longitudinal records are checkpoint aggregates without case identifiers, so they support within-stream comparison of the three regimes rather than paired tests or interval estimates.

The evaluation split is balanced by difficulty, but individual hard-constraint types overlap within a query. Table~\ref{tab:travelplanner_constraints} reports the resulting distribution.

\begin{table}[H]
\centering
\caption{TravelPlanner controlled-evaluation composition (180 queries). Difficulty is defined by the number of explicit hard constraints.}
\label{tab:travelplanner_constraints}
\small
\begin{tabular}{lrr@{\qquad}lrr}
\toprule
Difficulty & Count & Ratio & Constraint type & Count & Ratio \\
\midrule
Easy (0 constraints) & 60 & 33.33\% & House rule & 77 & 42.78\% \\
Medium (1 constraint) & 60 & 33.33\% & Room type & 64 & 35.56\% \\
Hard (3 constraints) & 60 & 33.33\% & Transportation & 51 & 28.33\% \\
& & & Cuisine & 48 & 26.67\% \\
\bottomrule
\end{tabular}
\end{table}

\paragraph{SpreadsheetBench and SkillCraft.}
We use a 400-case verified subset derived from the public SpreadsheetBench release~\cite{ma2024spreadsheetbench}, partitioned into 80/40/280 train/development/test cases with disjoint item and workbook identities. The held-out set contains 193 cell-level and 87 sheet-level cases. Success requires the executed workbook to match the verified target rather than merely producing plausible code. The paired admission study holds the model, split, optimizer, candidate stream, seed, and worker configuration fixed. For SkillCraft~\cite{chen2026skillcraft}, we regroup the public tasks by base family and use 11/4/6 disjoint train/development/test families (66/24/36 instances), preventing near-duplicate templates from crossing splits.

\subsection{Implementation and Reporting Details}\label{app:implementation_details}

All experiments use GPT-5.6-terra. The same configured backbone is used across methods and for target execution, trajectory analysis, and type-specific generation, with a fixed decoding configuration within each condition. Rules are injected into the parent context; Skills and Subagents are installed as directory-scoped modules using the host runtime's native convention. Subagents execute with their declared inputs and isolated working state. Artifact loading follows these existing runtime conventions; retrieval algorithms, directory scanning, and call routing are not experimental variables.

Reported success rates are consistent with integer case counts on the held-out sizes stated above: 134 for ALFWorld, 144 for ScienceWorld, 180 for TravelPlanner, 36 for SkillCraft, and 280 for SpreadsheetBench. SkillCraft therefore resolves success only in steps of 2.78 points, which is why several baselines share a value in its row.

Task success is the primary metric. The main turn statistic counts top-level parent-agent responses and excludes tool-return messages, nested Subagent responses, and offline artifact construction or validation. The \emph{Online Compute Accounting} section below reports nested responses and expanded actor counts separately. Percentage-point deltas are always absolute. When two conditions provide case-level binary outcomes on the same examples, we use a two-sided exact McNemar test over discordant pairs.

The causal controls differ by research question. RQ2 type-removal conditions mask one type from the same frozen repository; compiler ablations rebuild from the same evidence while changing only the selection policy; and validation ablations hold candidate generation fixed while changing admission. RQ3 gives every method the same ordered learning stream and evaluates every 20 tasks on a fixed held-out set. RQ4 freezes each source repository and prohibits target-side updates.

\subsection{Compiler Instantiation and Prompt Structure}\label{app:compiler_instantiation}

The analyzer and the three type-specific generators are structured-output LLM modules rather than trained classifiers. The analyzer prompt contains (i) completed trajectories with stable record and span identifiers, tool observations, and terminal evaluator feedback; (ii) successful trajectories from the same regime as preservation evidence; (iii) instructions to separate observed facts from proposed corrections; and (iv) a schema for activation, behavioral scope, persistent state, dependent decisions, and completion. Every generated field must cite one or more input span identifiers. Unparseable output, a missing required field, or an invalid citation defers the record rather than triggering a best-effort repair.

Grouping is criterion-based rather than embedding clustering or $k$-means. For record $r_i$, the analyzer emits an evidence-linked descriptor $q_i=(\alpha_i,\beta_i,\mathcal{S}_i,\mathcal{Q}_i,\gamma_i)$. Define pairwise compatibility by
\begin{equation}
g(i,j)=\mathbf{1}[\alpha_i\approx_A\alpha_j]\,
\mathbf{1}[\beta_i\approx_B\beta_j]\,
\mathbf{1}[\mathcal{S}_i\bowtie_S\mathcal{S}_j]\,
\mathbf{1}[(\mathcal{Q}_i,\gamma_i)\bowtie_C(\mathcal{Q}_j,\gamma_j)].
\label{eq:app_grouping}
\end{equation}
The structured analyzer judges whether activation and corrective objective have the same observable meaning ($\approx_A,\approx_B$) and whether state, decisions, and completion are non-conflicting ($\bowtie_S,\bowtie_C$), citing the spans that justify each judgment. A proposed cluster is accepted only when every pair satisfies $g(i,j)=1$ and every citation resolves. These are operational compatibility predicates, not assumed transitive equivalence relations. A successful record with the same activation but incompatible prescribed behavior blocks the merge and becomes a negative-boundary case. The deterministic controller checks pairwise membership, citation validity, required fields, and explicit conflicts; it does not invent a missing trigger or resolve a semantic disagreement. For the frozen RQ1/RQ2 repositories, the compilation pool is the complete learning partition in Table~\ref{tab:benchmark_protocols}. The longitudinal study exposes successive 20-task learning blocks at the reported checkpoints. Held-out evaluation records never enter either pool.

For each attempted schema, the generator returns both an artifact candidate and a four-field closure report. Each observation, state, decision, and completion field contains a Boolean status, cited evidence, and either the owning contract slot or an insufficiency witness. The semantic proposal is LLM-generated, but the branch decision is mechanical: a schema closes only if all four statuses are true, all required contract slots are present, and every citation resolves to input evidence. Otherwise the controller carries the explicit witness to the next wider schema. The contract gate subsequently combines deterministic schema and reference checks with representation-specific execution tests; the replay gate uses the task evaluator on correction and preservation cases rather than an LLM preference judgment. Language-model generation supplies the semantic candidates, while repository writes stay under explicit, inspectable conditions.

\subsection{Statistical Interpretation}\label{app:statistical_interpretation}

Comparisons in this paper are paired at the case level. Each controlled condition is executed once under the shared fixed protocol, and every reported difference is measured on the same cases with the same evaluator and execution budget, so the evidence rests on per-case outcomes rather than on run-level averages. Exact McNemar tests over discordant pairs quantify controlled differences, and Wilson intervals quantify finite case-set uncertainty: TravelPlanner's 145 successes among 180 cases correspond to a 95\% interval of [74.16, 85.67]. Because repeated-run variance is not estimated, we restrict conclusions to effects that are large relative to this uncertainty, and we report the 1.11-point Rule delta as statistically indistinguishable ($p=.875$) rather than as a positive result.

\subsection{Online Compute Accounting}\label{app:online_compute_accounting}

The TravelPlanner archive contains per-case parent messages, Subagent execution records, LLM-call ledgers, token usage, and task duration. We define $T_{\mathrm{parent}}$ as successful top-level responses (the statistic used in the main-paper results table), $N_{\mathrm{invoke}}$ as parent responses that call \texttt{ask\_subagent}, and $T_{\mathrm{sub}}$ as successful Subagent-internal responses. The expanded actor count is $T_{\mathrm{parent}}+T_{\mathrm{sub}}$; it retains the main table's exclusion of model-backed Planner/parser tool stages, whose calls are included separately in the all-stage accounting below. Table~\ref{tab:travelplanner_compute_accounting} reports the resulting per-case means.

\begin{table}[H]
\centering
\caption{TravelPlanner delegation accounting from 180 case-level traces. Except for Pass, entries are means per case. ``Actor total'' is $T_{\mathrm{parent}}+T_{\mathrm{sub}}$.}
\label{tab:travelplanner_compute_accounting}
\small
\setlength{\tabcolsep}{4.5pt}
\begin{tabular}{lrrrrr}
\toprule
Condition & Pass & Parent & Invoke & Sub. resp. & Actor total \\
\midrule
Full AutoRefine & 80.56\% & 20.68 & 1.48 & 8.79 & 29.47 \\
w/o Subagent & 72.22\% & 34.03 & 0.00 & 0.00 & 34.03 \\
\bottomrule
\end{tabular}
\end{table}

The raw Full counts are 3,722 parent responses, 266 invocations, and 1,583 successful internal responses. Calls occur in 179/180 cases: one case makes no call, 92 make one, and 87 make two. All 266 runs complete and return a summary, with a mean internal depth of 5.95 responses per invocation. The call ledger additionally records 53 failed Subagent attempts that are retried; as failures do not yield an assistant response, they are excluded from $T_{\mathrm{sub}}$ but included in API-attempt and latency totals.

Across all LLM stages, Full makes 5,765 attempts and receives 5,607 successful responses (32.03 and 31.15 per case), versus 6,562 and 6,351 (36.46 and 35.28) without Subagents. Full consumes 58.27M tokens (323.72K per case), of which the Subagent stage accounts for 17.81M (98.94K per case); the removal consumes 61.34M (340.78K per case). Delegation reduces parent responses by 13.35, expanded actor responses by 4.56, and tokens by 5.00\% in this controlled run. It nevertheless raises mean task wall time from 98.36s to 126.08s (28.18\%), exposing a latency cost that turn counts alone conceal.

These quantities separate two comparisons that should not be conflated. Relative to AutoManual's 15.4 in the main-paper results table, AutoRefine's 20.68 is already a parent-only count: 1.48 of those parent turns directly invoke a Subagent, while the 8.79 internal responses are additional. Relative to the controlled w/o-Subagent condition, delegation improves pass rate and reduces both parent and expanded actor counts, but takes longer in wall-clock time. We therefore report turns jointly with tokens and latency.

\subsection{TravelPlanner Difficulty Breakdown}\label{app:additional_experiments}

Table~\ref{tab:travelplanner_difficulty} gives the exact counts underlying the artifact-availability rows of the main paper's ablation table. Skills have the broadest effect across all difficulty levels. Subagents contribute primarily on Medium and Hard tasks, where the plan exposes longer-lived comparison state and independently completable objectives. Rules have a targeted effect on Hard tasks, consistent with local guards becoming decisive only when several explicit constraints interact.

\begin{table}[H]
\centering
\caption{Pass rate by TravelPlanner difficulty (60 cases per bucket). Deltas are relative to Full within each bucket.}
\label{tab:travelplanner_difficulty}
\small
\setlength{\tabcolsep}{8pt}
\begin{tabular}{lrrrrrrr}
\toprule
& \multicolumn{2}{c}{Easy} & \multicolumn{2}{c}{Medium} & \multicolumn{2}{c}{Hard} & Overall \\
\cmidrule(lr){2-3}\cmidrule(lr){4-5}\cmidrule(lr){6-7}
Condition & Pass & $\Delta$ & Pass & $\Delta$ & Pass & $\Delta$ & Pass \\
\midrule
\textbf{Full AutoRefine} & \textbf{50/60} & -- & \textbf{47/60} & -- & \textbf{48/60} & -- & \textbf{80.56\%} \\
w/o Rule & 50/60 & 0.00 & 49/60 & +3.33 & 44/60 & -6.67 & 79.44\% \\
w/o Skill & 39/60 & -18.33 & 36/60 & -18.33 & 34/60 & -23.33 & 60.56\% \\
w/o Subagent & 48/60 & -3.33 & 41/60 & -10.00 & 41/60 & -11.67 & 72.22\% \\
\bottomrule
\end{tabular}
\end{table}

Paired against Full, removing Skills yields 46 Full-only versus 10 ablation-only successes ($p=1.25\times10^{-6}$); removing Subagents yields 31 versus 16 ($p=.040$); and removing Rules yields 21 versus 19 ($p=.875$). The aggregate Rule delta is therefore not statistically distinguishable in this run, whereas the Skill and Subagent effects are supported by paired outcomes.

\subsection{Exact Repository-Evolution Checkpoints}

\begin{figure}[H]
\centering
\includegraphics[width=0.86\textwidth]{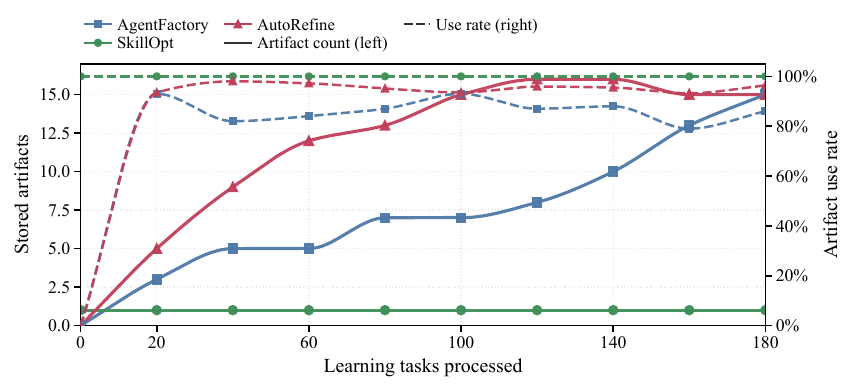}
\caption{Repository size (solid, left axis) and artifact use rate (dashed, right axis) over the same learning stream. AgentFactory keeps a high use rate while its held-out pass rate falls, so invocation frequency is not evidence of utility.}
\label{fig:rq3_size_use}
\end{figure}

Table~\ref{tab:rq3_checkpoints} reports the values plotted in the main paper's repository-dynamics figure, and Figure~\ref{fig:rq3_size_use} enlarges the size and use-rate panel of that figure. ``Signed degradation'' is a net statistic over the baseline-success ledger: positive values denote net loss and negative values denote net recovery. It does not imply that no individual case regressed. Artifact use is the fraction of stored artifacts invoked over the held-out evaluation. High use alone is not evidence of benefit, as shown by AgentFactory's simultaneous 86\% use and 55.0\% degradation at the final checkpoint.

\begin{table}[H]
\centering
\caption{Repository evolution on the common ordered TravelPlanner learning stream. Pass is held-out pass rate (\%), Degr.\ signed degradation (\%), and Use the fraction of stored artifacts invoked (\%). SkillOpt maintains one rewritten Skill throughout, so its size is 1 and its use rate 100\% at every checkpoint. R/S/A gives AutoRefine's Rule/Skill/Subagent counts. The zero-task row is the shared artifact-free reference of 62/180 cases, so it is reported at count precision.}
\label{tab:rq3_checkpoints}
\scriptsize
\setlength{\tabcolsep}{4.2pt}
\begin{tabular}{rrrrrrrrrrr}
\toprule
& \multicolumn{4}{c}{AgentFactory} & \multicolumn{2}{c}{SkillOpt} & \multicolumn{4}{c}{AutoRefine} \\
\cmidrule(lr){2-5}\cmidrule(lr){6-7}\cmidrule(lr){8-11}
Learned tasks & Pass & Size & Degr. & Use & Pass & Degr. & Pass & R/S/A & Degr. & Use \\
\midrule
0   & 34.44 & 0  & 0.0  & --   & 34.44 & 0.0  & 34.44 & 0/0/0 & 0.0  & -- \\
20  & 17.0 & 3  & 37.1 & 93.0 & 58.0 & 21.4 & 65.0 & 3/2/0 & 0.0  & 93.5 \\
40  & 14.0 & 5  & 40.7 & 82.0 & 63.0 & 16.7 & 78.0 & 4/3/2 & -3.5 & 98.0 \\
60  & 20.0 & 5  & 23.6 & 84.0 & 64.0 & 17.5 & 90.0 & 6/4/2 & -5.9 & 97.2 \\
80  & 17.0 & 7  & 30.0 & 87.0 & 71.0 & 13.5 & 91.0 & 6/4/3 & -7.1 & 95.1 \\
100 & 23.0 & 7  & 30.0 & 93.0 & 72.0 & 10.3 & 90.0 & 8/4/3 & -5.9 & 93.5 \\
120 & 18.0 & 8  & 37.1 & 87.0 & 75.0 & 13.5 & 91.0 & 8/5/3 & -3.5 & 95.9 \\
140 & 23.0 & 10 & 30.0 & 88.0 & 73.0 & 15.1 & 91.0 & 8/5/3 & -3.5 & 95.5 \\
160 & 23.0 & 13 & 44.3 & 79.0 & 77.0 & 17.5 & 91.0 & 8/4/3 & -3.5 & 93.2 \\
180 & 14.0 & 15 & 55.0 & 86.0 & 69.0 & 19.0 & 89.0 & 8/4/3 & -4.7 & 96.4 \\
\bottomrule
\end{tabular}
\end{table}

AutoRefine reaches 90\% after 60 learning tasks and then remains between 89\% and 91\%. Its repository grows only when candidates pass admission and can shrink when a validated revision supersedes an older artifact; artifact count is therefore not the optimization target. In contrast, AgentFactory accumulates 15 artifacts while returning to 14\% success, and SkillOpt's final overwrite reduces success from 77\% to 69\%. These trajectories explain why the final checkpoint alone is insufficient to assess reliable accumulation.

\subsection{Admission Quality and Offline Cost}

The SpreadsheetBench comparison isolates the cost of validation-gated heterogeneous updates against the SkillOpt control under a shared model, split, candidate stream, and seed. AutoRefine improves 35 cases that the control misses, while the control improves 18 cases that AutoRefine misses, giving exact McNemar $p=.027$. AutoRefine's 73.57\% success has a Wilson 95\% interval of [68.11, 78.39]. Table~\ref{tab:spreadsheet_admission_cost} reports admission quality together with offline construction cost.

\begin{table}[H]
\centering
\caption{SpreadsheetBench admission quality and offline construction cost on 280 held-out cases. Tokens and wall time include optimization and validation, not only final execution.}
\label{tab:spreadsheet_admission_cost}
\small
\begin{tabular}{lrrrrr}
\toprule
Method & Overall & Cell-level & Sheet-level & Offline tokens & Wall time \\
\midrule
SkillOpt control & 67.50\% & 62.69\% & \textbf{78.16\%} & \textbf{3.34M} & \textbf{1.35 h} \\
\textbf{AutoRefine gate} & \textbf{73.57\%} & \textbf{72.02\%} & 77.01\% & 9.43M & 4.12 h \\
\midrule
Difference / ratio & +6.07 pp & +9.33 pp & -1.15 pp & 2.82$\times$ & 3.05$\times$ \\
\bottomrule
\end{tabular}
\end{table}

The gain is concentrated in cell-level manipulation; on sheet-level cases, AutoRefine solves one fewer item than the control (67/87 versus 68/87, a $-1.15$-point difference). The gate therefore improves overall admission quality in this setting but is not an efficiency result: it spends substantially more offline computation to test candidate contracts and replay obligations.

\subsection{SkillFlow Cross-Domain Transfer}\label{sec:skillflow_transfer}

SkillFlow~\cite{zhang2026skillflow} contains 166 executable tasks organized into 20 workflow families and five domains: Finance \& Economics (FE), Operations \& Supply Chain (OS), Healthcare \& Life Sciences (HLS), Governance \& Strategy (GS), and Data \& Document Intelligence (DDI). Domain-Agnostic Execution Flow preserves workflow topology while replacing domain-specific entities. Artifacts are learned only from source families, the repository is frozen, and target evaluation performs no writes.

\begin{figure}[H]
\centering
\includegraphics[width=0.92\columnwidth]{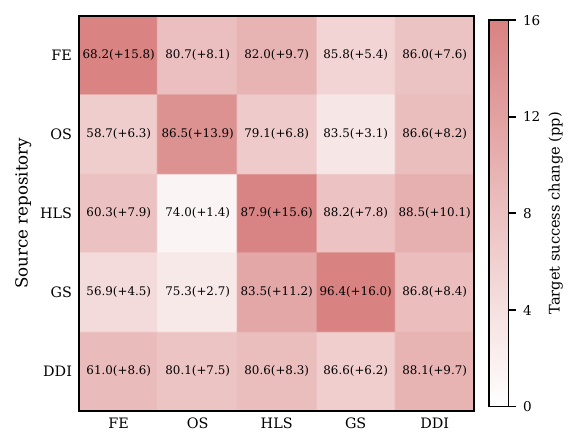}
\caption{SkillFlow transfer matrix. Cells show target pass rate and the change from the target-only baseline in percentage points; final rates are baseline plus gain.}
\label{fig:transfer_heatmap}
\end{figure}

The per-cell gains are reported in the main paper; Figure~\ref{fig:transfer_heatmap} adds the resulting absolute pass rates, which are always the target-only baseline plus the gain. Two properties of the matrix are worth separating. First, the off-diagonal gains are uniformly positive, which is what supports transfer of workflow structure rather than of domain vocabulary. Second, the matched-domain repository is the strongest source in four of the five columns, the exception being DDI, where the HLS repository transfers 10.1 points against the matched 9.7. Domain-specific evidence therefore carries an average advantage, not a per-column guarantee, and the mean matched-versus-cross gap (14.20 against 6.99 points) is the size of that average advantage. Averaged over source repositories, HLS and DDI are the most receptive targets (10.32 and 8.80 points) and OS the least (6.72); averaged over targets, FE is the strongest source (9.32), followed by HLS and GS (8.56 each). Per-cell denominators and independent repeats are not available in the retained records, so the matrix supports the direction and rough magnitude of transfer rather than per-cell significance.

\subsection{ALFWorld Heat-Then-Place Case Study}\label{sec:case_study_heat}

Heat-then-place tasks require the agent to find an object, acquire it, apply the required state change, deliver it to the requested receptacle, and stop after confirmed placement. The frozen repository represents the multi-stage search--acquire--heat--deliver routine as a Skill and the final stopping condition as a Rule. It does not introduce a Subagent because the parent already owns the environment state and task completion.

Across the 23 \texttt{valid\_unseen} heat-then-place tasks, the frozen Rule-and-Skill repository reduces mean environment steps from 18.91 to 11.48, a 39.3\% reduction. The example demonstrates the minimal-boundary principle in execution: a reusable multi-step procedure and a local termination guard suffice, so a delegated control loop would add ownership without closing an additional evidence-supported requirement.

\section{Limitations and Broader Impacts}

\subsection{Limitations}\label{app:limitations}

\paragraph{Heuristic boundary selection.}
The compiler is not trained against human type annotations, and the experiments evaluate downstream utility rather than classification accuracy. Closure is therefore a heuristic selection rule whose structural preconditions, contract slots and resolvable citations, are enforced mechanically, while the underlying semantic judgment is language-model proposed. An analyzer can still omit a required observation, state variable, dependent decision, or completion condition, and such an omission may cause a narrower artifact to appear closed. Because closure by itself never authorizes a write, the consequence of a wrong boundary is bounded by the contract and replay gates, whose decisions come from execution and the task evaluator; these gates cannot, however, prove that the intervention specification itself is complete.

\paragraph{Evidence-bounded validation.}
The replay gate establishes improvement and preservation only on cases linked to the candidate's evidence. It is not a global non-regression guarantee over the task distribution. Distribution shift, interactions among independently admitted artifacts, or an unobserved negative-boundary case can still produce interference. Revisions replay predecessor-associated cases, but the current method does not solve optimal global regression scheduling.

\paragraph{Runtime dependence.}
The Rule--Skill--Subagent distinction assumes a runtime that can inject local guards, install parent-executed modules, and invoke isolated workers with explicit I/O. Other runtimes may expose different state or tool boundaries. The operational definition transfers, but a contract may require adaptation before its isolation and termination semantics are equivalent.

\paragraph{Statistical scope and benchmark saturation.}
The \emph{Statistical Interpretation} section states the statistical scope: paired case-level tests and Wilson intervals over a fixed protocol, without an estimate of repeated-run variance. ALFWorld is nearly saturated and SkillCraft reaches a 100\% ceiling, limiting discrimination among strong methods; TravelPlanner and SpreadsheetBench provide the clearest separation in the present evaluation.

\paragraph{Delegation-cost attribution.}
The TravelPlanner ledgers permit actor-level decomposition, and the \emph{Online Compute Accounting} section reports parent responses, invocations, internal responses, tokens, and wall time. This supports a direct same-infrastructure comparison with the w/o-Subagent condition. As nested delegation introduces additional internal work, turn counts should be interpreted jointly with tokens and wall time.

\paragraph{Construction cost.}
Validation shifts computation from repeated online recovery to offline repository construction. On SpreadsheetBench, AutoRefine uses 2.82$\times$ the optimizer tokens and 3.05$\times$ the wall time of the SkillOpt control. This cost may be justified when artifacts are reused frequently, but the present experiments do not identify a universal break-even point.

\subsection{Broader Impacts}\label{app:broader_impacts}

Typed contracts and evidence-linked replay can make learned agent capabilities more inspectable: each artifact declares what it owns, when it runs, and which cases licensed its admission. This can reduce repeated trial-and-error and support targeted audits. However, a validated artifact can also propagate a mistaken or harmful strategy at scale, especially when its activation boundary extends beyond the replay evidence. High-impact deployment therefore requires access control, immutable provenance logs, repository scoping, and human review in addition to the gates studied here.

Source trajectories may contain personal, proprietary, or security-sensitive information. Systems that construct persistent artifacts should minimize and redact trajectories before extraction, restrict artifact visibility to the intended domain, and enforce retention and deletion policies. Validation checks behavioral utility; it does not by itself prevent privacy leakage, malicious capability accumulation, or misuse in domains such as fraud, spam, or social engineering.

\end{document}